\documentclass[review]{elsarticle}

\usepackage{lineno}
\modulolinenumbers[5]

\usepackage[utf8]{inputenc} 
\usepackage[T1]{fontenc}    
\usepackage{url}            
\usepackage{booktabs}       
\usepackage{amsfonts}       
\usepackage{nicefrac}       
\usepackage{microtype}      
\usepackage{multirow, multicol}
\usepackage{color}
\usepackage{amsmath}
\usepackage{microtype}      
\usepackage{enumerate}
\usepackage{graphicx}
\usepackage{comment}

\usepackage{enumitem}
\usepackage{dsfont}
\usepackage{wrapfig}
\usepackage{hyperref}       

\newcommand*{\eg}{\emph{e.g.,~}}
\newcommand*{\ie}{\emph{ie.,~}}
\newcommand*{\etal}{\emph{et al.~}}

\begin{document}

\begin{frontmatter}

\title{The Foes of Neural Network's Data Efficiency Among Unnecessary Input Dimensions}


\author[mit]{Vanessa D'Amario\corref{mycorrespondingauthor}}
\author[stanford]{Sanjana Srivastava}
\author[fujitsu]{Tomotake Sasaki}
\author[mit]{Xavier Boix\corref{mycorrespondingauthor}}

\address[mit]{Center for Brains, Minds \& Machines \\ Department of Brain and Cognitive Science, \\ Massachusetts Institute of Technology \\ 
Cambridge, MA 02139, USA}
\address[stanford]{Stanford University \\ Stanford, CA 94305, USA}
\address[fujitsu]{Artificial Intelligence Laboratory\\
  Fujitsu Limited \\
  Kawasaki, Kanagawa 211-8588, Japan}

\cortext[mycorrespondingauthor]{Corresponding authors:  {vanessad@mit.edu} and  {xboix@mit.edu}}

\begin{abstract}

Datasets often contain input dimensions that are unnecessary to predict the output label, \eg background in object recognition, which lead to more trainable parameters. Deep Neural Networks (DNNs) are robust to increasing the number of parameters in the hidden layers, but it is unclear whether this holds true for the input layer. In this letter, we investigate the impact of unnecessary input dimensions on a central issue of DNNs: their data efficiency, \ie the amount of examples needed to achieve certain generalization performance. Our results show that unnecessary input dimensions that are {task-unrelated}  substantially degrade data efficiency. This highlights the need for mechanisms that remove {task-unrelated} dimensions to enable data efficiency gains.

\end{abstract}


\end{frontmatter}


\section{Introduction}

The success of Deep Neural Networks (DNNs) contrasts with the still distant goal of learning with few training examples as in biological systems,~\ie in a data efficient manner~\cite{hassabis2017neuroscience}. Little is known about what aspects could improve the DNN's data efficiency.

DNNs are usually trained on high dimensional datasets (\eg images and text), and many input dimensions of the DNN may be unnecessary to predict the ground-truth label as they are unrelated and/or redundant to the task at hand. 
Classic machine learning theory predicts that unnecessary input dimensions may degrade the DNN's data efficiency~\cite{hastie2009elements}, as the classifier may overfit to the unnecessary input dimensions if not enough training examples are provided to learn to discard them.   

However, DNNs have challenged classic machine learning theories as they can achieve high test accuracy despite having a number of trainable parameters much larger than the number of training examples, \ie DNNs are overparameterized~\cite{zhang2016understanding, nakkiran2019deep}. Since unnecessary input dimensions lead to more overparameterization, it is unclear in what way DNNs suffer from unnecessary input dimensions and whether more data is needed to learn to discard them. 
In this letter, we show that the DNNs' data efficiency depends on whether the unnecessary dimensions are \emph{task-unrelated} or \emph{task-related} (redundant with respect to other input dimensions). Namely, increasing the number of \emph{task-unrelated} dimensions leads to a substantial drop of data efficiency, while increasing the number of \emph{task-related} dimensions  that are linear combinations of other \emph{task-related} dimensions, helps to alleviate the negative impact of the \emph{task-unrelated} dimensions.

\section{Related Works}

We now relate this work with the effect of background on the generalization abilities of DNNs in object recognition, and also with the DNNs generalization abilities depending on the number of parameters of the network.

    \subsection{Object's Background and DNN Generalization} 
    
    Previous works have shown that DNNs for image recognition fail to classify objects in novel and uncommon backgrounds~\cite{choi2012context,beery2018recognition,volokitin2017deep}.  Remarkably, popular object recognition datasets are biased to such an extent that DNNs can predict the object category even when the objects are removed from the image~\cite{zhu2016object, xiao2021noise}. 
    
    Barbu~\etal\cite{barbu2019objectnet} introduced a new benchmark which addresses the biased co-occurrence of objects and background, among other types of bias. DNNs exhibit large performance drops in this benchmark compared to ImageNet~\cite{ImageNet09}. Recently, Borji has shown that a large portion of the performance drop comes from the bias in object's background, as classifying the object in isolation substantially alleviates the performance drop \cite{borji2021contemplating}. 
    
    
    In contrast to previous works, we analyse the impact of the object's background to the DNN's generalization performance when the dataset is unbiased,~\ie the statistics of the object's background are the same between training and testing and there is no bias in the co-occurrence of object and background. To the best of our knowledge, our work is the first to investigate the effects of object's background on DNNs when the object's background is unbiased. 
    
\subsection{Overparameterization and Data Dimensionality}


A remarkable characteristic of DNNs is that the test error follows a double-descend when the DNN's width is increased by adding more hidden units. Thus, the test error decreases as  the network's width is increased in both the underparameterized and overparameterized regimes, except in a critical region between these two where a substantial error increase can take place~\cite{advani2017high,belkin2019reconciling,nakkiran2019deep}.
This phenomenon relates to unnecessary input dimensions because these also increase the width of the network, albeit in the input layer rather than in the intermediate layers.  
As we show in the sequel, increasing the number of unnecessary input dimensions can have the opposite effect of increasing the number of hidden units in the test error.

Another strand of research relates the structure of the dataset with the generalization ability of the network. Several works in statistical learning theory for kernel machines relate the spectrum of the dataset with the generalization performance~\cite{zhang2005learning}. For neural networks, \cite{recanatesi2019dimensionality,ansuini2019intrinsic} define the intrinsic dimensionality based on the dimension of the data manifold. These works analyze how the network reduces the intrinsic dimension across layers. Yet, these metrics based on manifolds do not provide insights about how specific aspects of the dataset,~\eg unnecessary dimensions, contribute to the intrinsic dimensionality.

\section{Unnecessary Input Dimensions and Data Efficiency}\label{sec:linear_model}
We aim at analyzing the effect of unnecessary input dimensions on the data efficiency of DNNs. 
Let $\mathbf{x}$ be a vector representing a data sample, and let $\mathbf{y}$ be the ground-truth label of $\mathbf{x}$. We define  $f(\mathbf{x})=\mathbf{y}$ as the target function of the learning problem. Also, we use $[\mathbf{x}; \mathbf{u}]$ to denote the data sample $\mathbf{x}$ with unnecessary input dimensions appended to it. The unnecessary dimensions do not affect the target function of the learning problem,~\ie $g([\mathbf{x}; \mathbf{u}]) = f(\mathbf{x}) = \mathbf{y}$, where $g$ is the target function of the learning problem with unnecessary input dimensions. Each sample can have a different set of dimensions that are unnecessary,~\eg one sample could be $[\mathbf{x}_1; \mathbf{u}_1]$ and another be $[\mathbf{u}_2; \mathbf{x}_2]$.  Note that this variability is present in object recognition because the dimensions representing the object's background are unnecessary and vary across data samples, as the object can be in different image locations.

We define two types of unnecessary input dimensions:  \emph{task-unrelated} and \emph{task-related}. Unnecessary input dimensions are \emph{task-unrelated}  when they are independent of $\mathbf{x}$,~\ie they can not be predicted from $\mathbf{x}$, as in unbiased object's background. Otherwise, the unnecessary dimensions are \emph{task-related}, which are equivalent to redundant dimensions. An example that leads to more \emph{task-related} unnecessary dimensions is upscaling the image.

To study the effect of unnecessary input dimensions, we measure the test accuracy of DNNs trained with different amounts of unnecessary input dimensions and training examples. 
Given a DNN architecture and a dataset with a fixed amount of unnecessary dimensions, we define the \emph{data efficiency} of the DNN as the 
Area Under the Test Curve (AUTC)
for the DNN trained with different number of training examples. The curve is monotonically increasing, as more training examples lead to higher test accuracy, and the AUTC measures the area under it. We normalize the AUTC to be between $0$ and $1$, where $1$ is the maximum achievable, and it corresponds to $100\%$ test accuracy for all number of training examples. In the experiments where the number of training examples spans several orders of magnitude, we calculate the AUTC by converting the number of training examples in logarithmic scale, such that all orders of magnitude are equally taken into account.

\section{Datasets and Networks}\label{sec:datasets}
We now introduce the datasets and networks we use in the experiments (refer to  \ref{app:details_data_generation} and \ref{app:networks_details} for additional details).  

\subsection{Linearly Separable Dataset} 
We use a linearly separable dataset for binary classification, as it facilitates relating results of classic machine learning  and DNNs.
We generate a binary classification dataset of $30$ input dimensions, which follow a Gaussian distribution with $(\mu=0, \sigma=1)$. The ground-truth label is the output of a linear classifier, such that the dataset is linearly separable with a hyperplane randomly chosen. Unnecessary input dimensions are appended to the data samples.  \emph{Task-unrelated} dimensions follow a Gaussian distribution with $(\mu=0,\sigma=0.1)$. The \emph{task-related} dimensions are linear combinations of the dimensions of the original dataset samples.

 We evaluate the following linear and Multi-Layer Perceptron (MLP) networks: linear network trained with square loss (pseudo-inverse solution), MLP with linear activation functions trained with either square loss or cross entropy loss, and MLP with ReLU  trained with cross entropy loss.

\subsection{Object Recognition Datasets}
\label{sec:object_methods}
We evaluate object recognition datasets based on extensions of the MNIST dataset and the   Stanford dogs dataset~\cite{KhoslaYaoJayadevaprakashFeiFei_FGVC2011}.

\paragraph{Synthetic and Natural MNIST}
We generate two datasets based on MNIST: the synthetic MNIST and the natural MNIST, which have synthetic and natural background, respectively. In both datasets, the MNIST digit is always at the center of the image and normalized between $0$ and $1$.  

In the synthetic MNIST dataset, the \emph{task-unrelated} dimensions are sampled from a Gaussian distribution with $(\mu=0, \sigma=0.2)$ and the \emph{task-related} dimensions are the result of upscaling the MNIST digit. We also combine \emph{task-related} and \emph{unrelated} dimensions by fixing the size of the image and changing the ratio of \emph{task-related} and \emph{unrelated} dimensions by upscaling the MNIST digit. 

In the natural MNIST dataset, the background is taken from the Places dataset~\cite{zhou2014learning}, as in \cite{volokitin2017deep}. The size of the image is constant across experiments ($256\times 256$ pixels), and the size of the MNIST digits determines the amount of \emph{task-related} and \emph{unrelated} dimensions. 

We use the MLP with ReLU and cross entropy loss, and also Convolutional Neural Networks (CNNs). The architecture of the {CNN} consists of three convolutional layers each with max-pooling, followed by two fully connected layers. 
Since the receptive field size of the {CNN} neurons may have an impact on the data efficiency, we evaluate different receptive field sizes.  We use a factor $r$ to scale the receptive field size, such that the convolution filter size is $(r\cdot 3)\times (r\cdot 3)$ and the pooling region size is $(r\cdot 2)\times (r\cdot 2)$. We experiment by either fixing $r$ to a constant value or adapting $r$ to the scale of the MNIST digit, such that the receptive fields of the neurons capture the same object region independently of the scale of the digit.

\paragraph{Stanford Dogs}
  Recall our analysis focuses on  unnecessary input dimensions that are unbiased. We use the Stanford dogs dataset~\cite{KhoslaYaoJayadevaprakashFeiFei_FGVC2011} as it is reasonable to assume that the bias between breeds of dogs and background is negligible. This dataset contains natural images ($227\times 227$ pixels) of dogs  at different image positions. The amount of \emph{task-unrelated} dimensions is determined by the dog size, which is different for each image.  To evaluate the effect of unnecessary input dimensions, we introduce the following five versions of the dataset. Case 1 corresponds to the original image. In {case 2}, we multiply by zero the pixels of the background, which 
reduces the variability of the \emph{task-unrelated} dimensions.
In {case 3}, the dog is centered in the image. In {case 4}, we fix the ratio of \emph{task-related/unrelated} dimensions by centering the dog and scaling it to half of the image size. In {case 5}, we remove the background by cropping and scaling the dog.

We use a ResNet-18~\cite{he2016deep}, following the standard pre-processing of the image used in ImageNet.

\section{Results}

\begin{figure*}
    \centering
    \begin{tabular}{c}
         \includegraphics[width=1\textwidth]{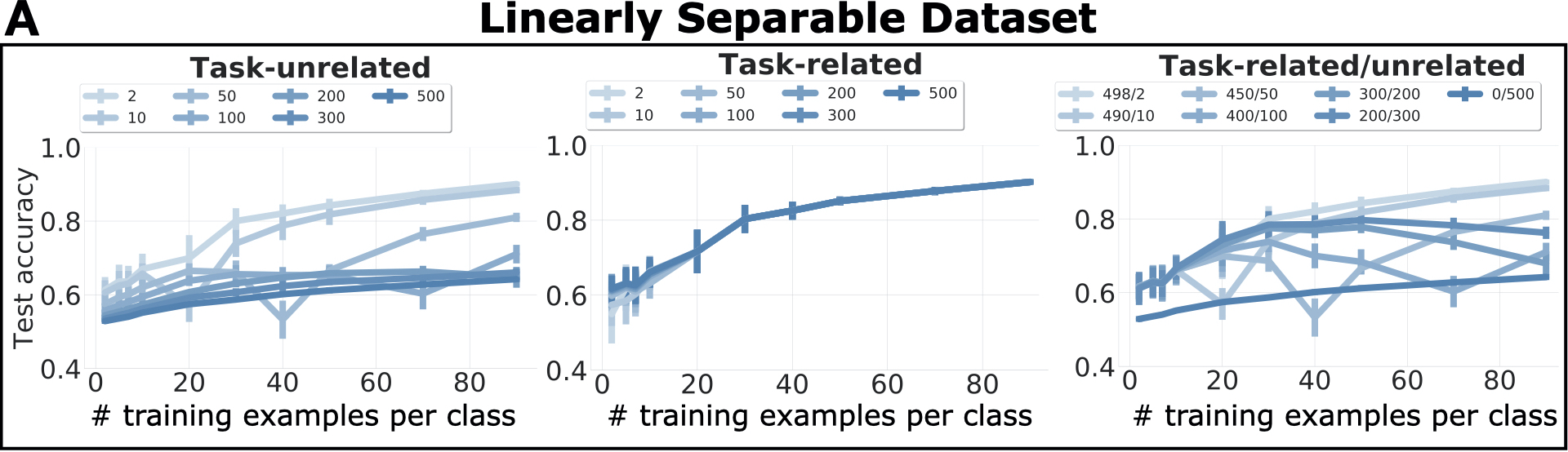} \\
        \includegraphics[width=1\linewidth]{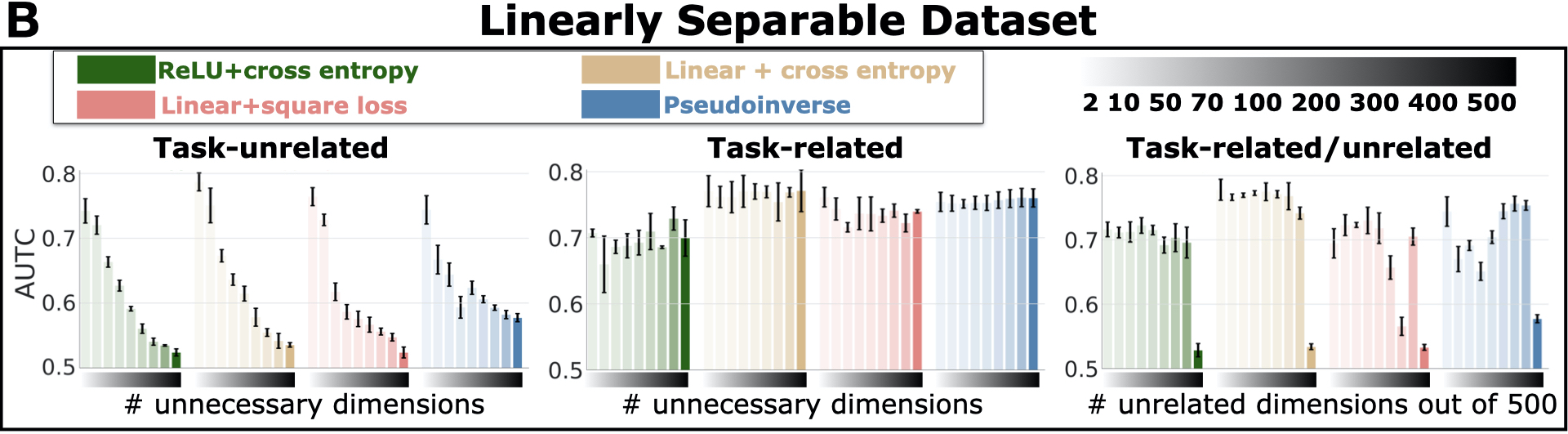} \\
        \includegraphics[width=1\linewidth]{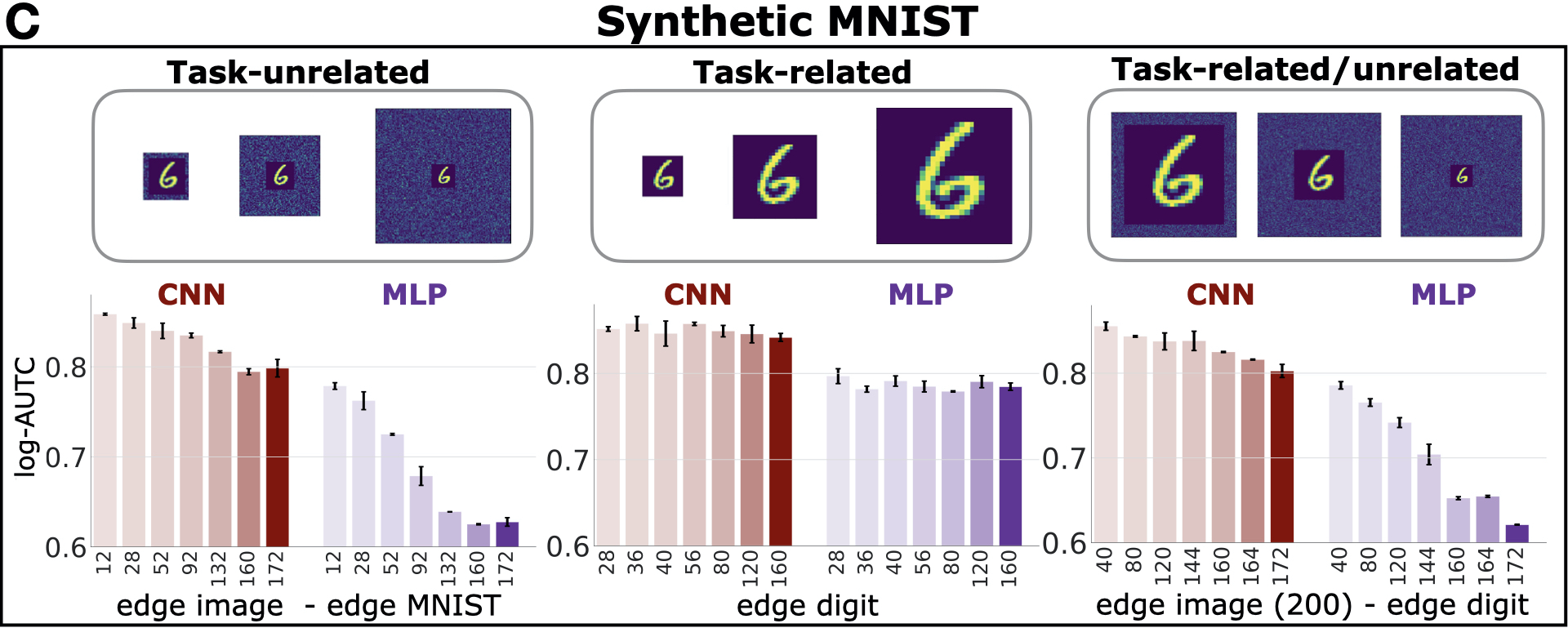} \\
        \includegraphics[width=1\linewidth]{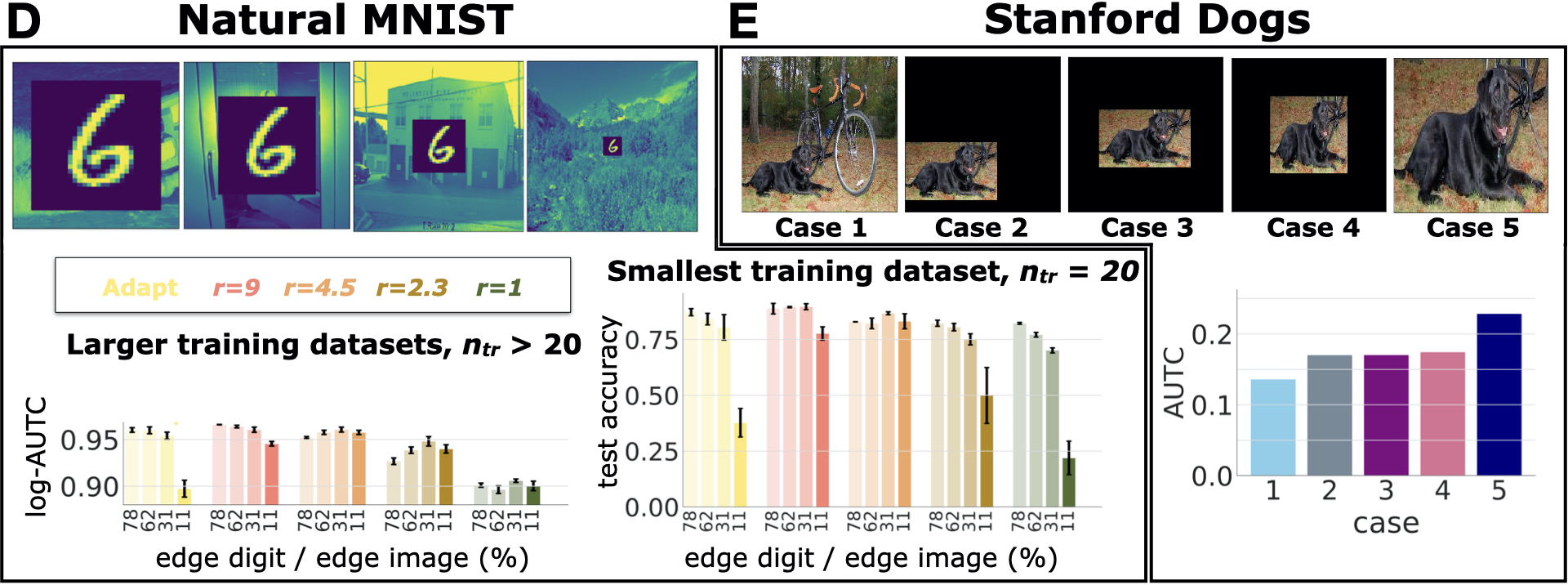}
    \end{tabular}
    \caption{Results of DNNs' data efficiency for different amount of unnecessary input dimensions,  tasks and networks. Error bars indicate standard deviation across experiment repetitions.}
    \label{fig:results_all}
\end{figure*}

All results are reported in Figure~\ref{fig:results_all}.

\subsection{Linearly Separable Dataset}

Figure~\ref{fig:results_all}A shows the test accuracy of the pseudo-inverse solution for different number of training examples and  unnecessary input dimensions.
Figure~\ref{fig:results_all}B reports the data efficiency of all networks tested for different number of unnecessary input dimensions. Recall that the data efficiency is measured with the AUTC and summarizes the test accuracy as a function of the amount of training examples,~\eg for the pseudo-inverse solution, the curves in Figure~\ref{fig:results_all}A are summarized by the AUTC in Figure~\ref{fig:results_all}B. We observe that increasing the  amount of \emph{task-unrelated} dimensions harms the data efficiency,~\ie the AUTC drops. Also, the \emph{task-related} dimensions alone do not harm data efficiency, and they alleviate the effect of the \emph{task-unrelated} dimensions. 

These results clarify the difference between robustness to overparameterization in intermediate layers and unnecessary input dimensions. Note that the effect on the test accuracy of increasing the number of hidden units is the opposite of increasing the number of \emph{task-unrelated} input dimensions,~\ie DNNs are not robust to all kinds of overparameterization.

Thus, the results on \emph{task-unrelated} dimensions can be predicted with classic machine learning results. In~\ref{sec:exact_solution}, we outline such results for the pseudo-inverse solution with \emph{task-unrelated} dimensions that are Gaussian-distributed. There, we show that \emph{task-unrelated} dimensions leads to the pseudo-inverse, Tihkonov-regularized solution calculated in the dataset without \emph{task-unrelated} dimensions. Since in this case the regularization can not be tuned or switched off as it is fixed by the number of \emph{task-unrelated} dimensions, it is likely to harm the test accuracy, as we have observed. 

The regularizer is beneficial in some specific cases. \ref{sec:real_regression_problem} highlights a regression problem in which certain amounts of \emph{task-unrelated} dimensions help to improve generalization. In Figure~\ref{fig:results_all}A, we  can also see that for a given number of training examples, increasing the number of \emph{task-unrelated} dimensions improves the test accuracy in some cases. This specific trends relate to the aforementioned double descend of DNNs~\cite{advani2017high, belkin2019reconciling}. As shown in~\cite{nakkiran2019deep}, the location of the critical region is affected by the number of training examples and the complexity of the model. Here, the complexity of the model is affected by the number of \emph{task-unrelated} dimensions due to its regularization effect. 

Finally, in \ref{app:additional_noise_non_separable} we show that the same conclusions are obtained for non-linearly separable datasets and other types of noise.

\subsection{Object Recognition Datasets}

Figure~\ref{fig:results_all}C shows the log-AUTC for the MLP and the CNN for different amount of unnecessary dimensions (an increasing amount as we move left to right), for the synthetic MNIST dataset.  In \ref{app:supplement_synthetic_mnist}, we also report the test accuracy for different number of unnecessary dimensions.  Conclusions are consistent with the previous results in the linearly separable dataset. Also, we observe that CNNs are overall much more data efficient than MLPs, which is expected because of their more adequate inductive bias given by the weight sharing of the convolutions.

Figure~\ref{fig:results_all}D shows results in natural MNIST dataset for different ratios of \emph{task-related/unrelated} dimensions. The plots compare CNNs with different receptive field sizes, represented by the factor $r$ (see section~\ref{sec:object_methods}).
Since the CNN achieves high accuracy with few examples, the mean and standard deviation of the log-AUTC (left plot) hardly shows any variation when computed on more than $20$ training examples per class. Yet, the gap of the testing accuracy is considerable for $20$ training examples per class (right plot). These results confirm that \emph{task-unrelated} dimensions degrades data efficiency independently of the receptive field sizes (see~\ref{app:supplement_natural_mnist} for additional results further supporting these conclusions).

Figure~\ref{fig:results_all}E shows results on the Stanford dogs dataset, namely the AUTC score across the five cases of unnecessary dimensions that we evaluate. This dataset serves to assess a more realistic scenario, where the objects can appear at different positions and scales. We observe that the \emph{task-unrelated} dimensions, which come from the background, harm the data efficiency (cases 1 to 4 versus case 5). Putting to zero the unnecessary dimensions improves the data efficiency of models trained on the original dataset (cases 2 to 4 versus case 1). This is expected as the \emph{task-unrelated} dimensions become redundant as they all take the same value in all images. We also observe that removing the variability of the position and scale of the object hardly affects the data efficiency (case 2 to 4). Thus, learning to discard the background requires more training examples than learning to handle the variability in scale and position of the object.

\section{Conclusions}
We have analyzed the effect of unnecessary input dimensions (\eg object's background). We found that \emph{task-unrelated} dimensions harm the data efficiency, while increasing the number of \emph{task-related} dimensions that are linear combinations of other \emph{task-related} dimensions help to alleviate the negative effect of \emph{task-unrelated} dimensions. These results demonstrate that the robustness of DNNs to overparameterization  is limited, as increasing the number of \emph{task-unrelated} input dimensions is a form of overparameterization that degrades the accuracy. 
Also, our results add to the growing body of works in object recognition that shows that bias in the object's background can undermine the reliability of DNNs. Here we have shown that the problem runs far deeper, as the object's background negatively affects the network even when there is no bias.  

Overall, these results suggest that  data efficiency gains could be enabled by mechanisms that remove \emph{task-unrelated} dimensions, such as foveation for image classification~\cite{ICLR16, akbas2017object}, or by adapting to DNNs regularization techniques that encourage predictions from a sparse subset of input dimensions (\eg $\ell_1$ regularization for linear regression~\cite{hastie2019statistical}).

\section*{Acknowledgment}
We would like to thank Pawan Sinha and Tomaso Poggio for useful discussions and insightful advice provided during this project. This work has been supported by the Center for Brains, Minds and Machines (funded by NSF STC award CCF-1231216), XB by the R01EY020517 grant from the National Eye Institute (NIH) and XB and VD by Fujitsu Laboratories Ltd. (Contract No. 40008819) and the MIT-Sensetime Alliance on Artificial Intelligence.
\bibliographystyle{elsarticle-num}

\bibliography{biblio.bib}
\appendix
\section{Data Generation Details}\label{app:details_data_generation}
We provide further details about the data generation. In \ref{app:details_lin_sep}, we discuss the linearly separable dataset, and \ref{app:object_classification_dataset} divides in three parts, where we introduce the synthetic MNIST, the natural MNIST, and the Stanford dogs datasets. 

\subsection{Linearly Separable Dataset}\label{app:details_lin_sep}
For the binary classification datasets, we generate the samples using a linear teacher network $\mathbf{y}_i = {W}^* \mathbf{x}_i $, with ${W}^*\in\mathbb{R}^{2\times 30}$ (\ie $o=2$, $p=30$) components, generated from a standard Gaussian distribution $(\mu=0, \sigma=1)$. Each input sample dimension is also drawn from a standard Gaussian distribution. To generate the labels, we quantize the output to two categories.
The \emph{task-unrelated} dimensions have $\sigma=0.1$, while \emph{task-related} dimensions are linear combinations of the $p$ (= 30) dimensions of the dataset, where the weights of the combinations are selected from a standard Gaussian distribution.

\subsection{Object Recognition Datasets}\label{app:object_classification_dataset}

\paragraph{Synthetic MNIST} In Figure \ref{fig:results_all}C, we show the different experiments on the Synthetic MNIST datasets. In all of these, the MNIST digit is normalized to have values between $0$ and $1$. 
For the experiments with  \emph{task-unrelated} dimensions, the image size increases as we add pixels with random values at the edge, which constitute the \emph{task-unrelated} dimensions. These random values are the absolute value of numbers extracted from a zero-mean Gaussian distribution with $\sigma=0.2$. The MNIST digit edge has size $28$ across all the \emph{task-unrelated} versions for this dataset, while the image edge is equal to $[40, 56, 80, 120, 160, 188, 200]$. For the experiment with \emph{task-related} dimensions, we upscale the MNIST digit using bi-cubic interpolation. After the upscale, the image size is equal to $[28, 36, 40, 56, 80, 120, 160]$. For the experiment with \emph{task-related/unrelated} dimensions, we use the same aforementioned procedure to generate the \emph{task-unrelated} and \emph{task-related} dimensions. We fix the image size to $200 \times 200$ pixels, while the upscaled MNIST digit edge is equal to $[28, 36, 40, 56, 80, 120, 160]$. The amount of training examples per class for all the experiments is equivalent to 
$n_{tr}=[1, 2, 5, 10, 20, 50, 100, 300, 500, 1000]$. 

\paragraph{Natural MNIST} We embed the MNIST digits on a natural background from the PLACES dataset \cite{zhou2014learning}, as in \cite{volokitin2017deep}. We generate the \emph{task-related/unrelated} by sampling without replacement a normalized natural image of size $256\times 256$ pixels. We superimpose at its center an upscaled and normalized digit from MNIST. The upscaled MNIST digit assumes sizes $[200, 150, 80, 28]$ as shown in Figure \ref{fig:results_all}D.  The number of examples per class corresponds to $n_{tr} = [20, 50, 100, 200, 300, 500, 1000]$.

\paragraph{Stanford Dogs}
All the images from the Stanford Dogs dataset, for all the five cases considered in Figure \ref{fig:results_all}E, are resized to $227\times 227$ pixels. We subtract to the RGB channels the values $R_{\text{mean}}=123.68$, $G_{\text{mean}}=116.78$, and $B_{\text{mean}}=103.94$, as done for recentering the ImageNet dataset.
We split the original training set into a training and a validation set, the former consisting of $90$ examples per class (breed), the latter consisting of $10$ examples per class. The amount of training examples per class used to compute the AUTC is $n_{tr} = [23, 45, 90]$.

\section{Architecture and Optimization Details}\label{app:networks_details}
In \ref{app:pseudo_inv_details}, we report the optimization details of the solution to the linear problem, using the pseudo-inverse. 
Architectures and optimization protocols of networks used on the linearly separable classes and the MNIST datasets are respectively in \ref{app:mlp_details} and \ref{app:mnist_details}. \ref{app:stanford} contains the optimization procedure used on the Stanford Dogs dataset.

\subsection{Linear classifier (pseudo-inverse)}\label{app:pseudo_inv_details}
 We repeat the classification experiments $10$ times. We use the pseudo-inverse solution as described in \eqref{sec:exact_solution} below.

\subsection{MLP}\label{app:mlp_details}
\paragraph{Architectures}
We consider MLPs with one hidden layer consisting of 128 nodes and a soft-max operation preceding the output. 


\paragraph{Optimization Protocol}
MLPs trained on the linearly separable binary classification task share the same optimization protocol. We fix the maximum number of epochs at 100. The convergence criterion is based on early stopping, with tolerance value $10^{-4}$ on the validation loss and patience of 8 epochs. Optimal batch size and learning rate are selected through a grid search procedure, respectively, over the arrays $[ 2, 10, 32, 50]$ and $[10^{-5}, 10^{-4}, 10^{-3}, 10^{-2}]$. We eliminate batch size values from the hyper-parameters search anytime those are equal to or larger than the number of training examples. We apply a reduction factor $1/10$ on the learning rate as the validation loss shows variations smaller than $10^{-4}$ over five epochs. The  initialization of the weights adopted across networks and layers is the Glorot uniform,~\ie uniform distribution in the interval $[-u, u]$, where $u = \sqrt{{6}/(\text{fan}_{\rm in} + \text{fan}_{\rm out})}$, with $\text{fan}_{\rm in}$ and $\text{fan}_{\rm out}$ respectively number of input and output units. We report mean and standard deviation of the AUTC values across three repetitions of each experiment.
The optimization protocol of MLP for MNIST is the same as the CNN detailed next.

\subsection{CNNs on synthetic MNIST and natural MNIST}\label{app:mnist_details}
\paragraph{Architectures}
The CNNs consist of three convolutional layers (iC-iiiC) followed by a flatten operation and two fully connected layers (iL and iiL). In Table~\ref{tab:architecture_cnn} we report: at the first row the number of filters or nodes, depending on the layer; at the second row, the size of filters and max-pooling operations; and at the last row, the non-linearity used, if any. The filters and max-pool operations have square dimensions, and their sizes at the first layers vary across experiments.
\begin{table}[!ht]
    \centering
    \caption{CNN for object recognition. From the top: number of filters *C or nodes *L at each hidden layer, filters size and max pooling sizes for *C;  activation functions following *C or *L. 
    }
    \begin{tabular}{|p{2.5cm}|p{1.4cm}|p{1.4cm}|p{1.4cm}|p{1.4cm}|p{1.4cm}|}
        \hline
                        &iC &iiC &iiiC &iL &iiL \\
        \hline
         \#filters/\#nodes & 32 & 64 & 64 & 64 & 10\\
         \hline
         filter, pooling &var, var & 3, 2 &3, none &none &none \\
         \hline
         act. function & ReLU &ReLU &None &ReLU &soft-max  \\
         \hline
    \end{tabular} 
    \label{tab:architecture_cnn}
\end{table}
In Table~\ref{tab:filter_size}, we report filter and max-pooling sizes corresponding to different receptive fields at the first layer. The AUTC and test accuracy for these filter/max-pooling configurations are evaluated on natural MNIST, in Figure~\ref{fig:results_all}D.
\begin{table}[!t]
    \centering
    \caption{Filters and max-Pooling sizes at the first layer for CNNs trained on Natural MNIST.}
    \begin{tabular}{|p{1.5cm}|p{1.2cm}|p{1.2cm}|p{1.2cm}|p{1.2cm}|p{1.2cm}|}
    \hline
         & Adapt & $r=9$ & $r=4.5$ & $r=2.3$  & $r=1$ \\
        \hline
         filter &$r\cdot3$ & 27 & 14 & 7 & 3  \\
         \hline
         pooling &$r\cdot 2$ & 18 & 9 & 5 & 2  \\
         \hline
    \end{tabular} 
    \label{tab:filter_size}
\end{table}

\paragraph{Optimization Protocols}
CNNs on synthetic MNIST share the same optimization protocol of MLPs. MLPs are provided with a vectorized version of the image. Models are optimized using stochastic gradient descent with null momentum. Batch size and learning rate are selected using a grid search procedure. The learning rate array has values $[1, 8 \cdot 10^{-1}, 5 \cdot 10^{-1}, 2\cdot 10^{-1}, 10^{-1}, 10^{-2}, 10^{-3}, 10^{-4}, 10^{-5}, 5\cdot 10^{-6}, 10^{-6}]$, the batch size array is $[10, 32, 50, 100]$. We eliminate batch size values from the hyper-parameters search anytime those are equal to or larger than the number of training examples. 
The convergence criterion is based on early stopping, convergence is reached when the validation loss shows across 10 repetitions variations smaller than  $10^{-6}$. The maximum amount of epochs is fixed at $500$. We reduce the learning rate by a factor $1/10$ when variations of the validation loss are smaller than $10^{-4}$ across $5$ epochs.
The initialization of the weights values across architectures is based on the Glorot uniform distribution, or uniform distribution in the interval $[-u, u]$, where $u = \sqrt{{6}/(\text{fan}_{\rm in} + \text{fan}_{\rm out})}$, with $\text{fan}_{\rm in}$ and $\text{fan}_{\rm out}$ respectively number of input and output units.

The natural MNIST optimization protocol shares most of its hyper-parameters with the synthetic MNIST case. The main variation is a different grid of learning rates, fixed at $[5 \cdot 10^{-1}, 2\cdot 10^{-1}, 10^{-1}, 10^{-2}, 10^{-3}, 10^{-4}]$. 

We report mean and standard deviation of log-AUTC and test accuracy across two repetitions of the experiments of synthetic MNIST, and three repetitions on natural MNIST.

\subsection{ResNet-18 on Stanford Dogs}\label{app:stanford}
\paragraph{Optimization}
Across all the experiments, the learning rate is equal to $1.28\cdot 10^{-1}$ at the first iteration. We then divide it by a factor of $10$ every time we reach a plateau of the validation accuracy. All the models are trained until we reach the plateau for the smallest learning rate considered, which is $1.28 \cdot 10^{-3}$. In all experiments, we observed no improvement of the validation accuracy after the plateau at learning rate $1.28\cdot 10^{-2}$.

\section{Theoretical Analysis of Linear Models with Unnecessary Input Dimensions}\label{app:theory}

In this section, we analyze linear, shallow networks trained with the square loss. In \ref{sec:exact_solution}, we introduce the theoretical analysis and in  \ref{sec:real_regression_problem}, we report experiments to illustrate the theoretical results. 

\subsection{Pseudo-inverse solution for linear, shallow networks}\label{sec:exact_solution}

We use $(X, Y)$ to denote a dataset where $X=[\mathbf{x}_1 \cdots  \mathbf{x}_n] \in\mathbb{R}^{p\times n}$ contains $n$ independent observations of $p$ uncorrelated features, and  $Y = [\mathbf{y}_1 \cdots \mathbf{y}_n] \in\mathbb{R}^{o\times n}$ denotes the respective ground-truth output. We define $p$ as the minimal dimensionality. We aim at estimating the function $f$, such that, for every example $(\mathbf{x}_i, \mathbf{y}_i)$, $\mathbf{y}_i = f(\mathbf{{x}}_i)$ holds approximately. The choice of a linear network $\hat{f}(\mathbf{x})=W\mathbf{x}$, with $W\in\mathbb{R}^{o\times p}$ and the square loss as cost function assures that, even for $p>n$ (overparameterized case), optimization through gradient descent leads to a unique solution that corresponds, among all, to the one with minimum norm, namely the pseudo-inverse \cite{scholkopf2001generalized}:  
\begin{equation}\label{eq:pseudoinverse_illposed}
	W^+ = Y (X^{\top} X)^{-1}X^{\top},\text { with } p > n.
\end{equation}

In the following, we introduce the theoretical solution to the linear model when the dataset contains \emph{task-related}, \emph{task-unrelated} and \emph{task-related/unrelated} dimensions. We assume that the model operates in the overparameterized regime, $p>n$, which is the most common in practice.


\paragraph{Task-related dimensions}
We assume that the unnecessary \emph{task-related} dimensions come from a linear transformation $T\in\mathbb{R}^{d \times p}$ of the vector of $p$ minimal dimensions. Thus, the dataset with unnecessary dimensions is the result of the transformation 
$F = [\mathbb{I}_p^{\top} \ \ T^{\top}]^{\top} \in \mathbb{R}^{(p+d)\times p}$, 
where $\mathbb{I}_p$ denotes the identity matrix of size $p$.
The pseudo-inverse solution of the linear model in Eq.~\eqref{eq:pseudoinverse_illposed} becomes the following:
\begin{equation}\label{eq:redundant_illposed_pinv}
  {W}^+ = Y (X^{\top} F^{\top} F X)^{-1} X^{\top} F^{\top}.
 \end{equation}
We further assume $F$ to be a tight frame~\cite{daubechies1992ten},~\ie  there is a unique scaling factor $a>0$ such that $\|F\mathbf{v}\|^2 = a\|\mathbf{v}\|^2$ holds for any vector $\mathbf{v}\in\mathbb{R}^p$. 
Then, Eq.~\eqref{eq:redundant_illposed_pinv} corresponds to ${W}^+=a^{-1} Y(X^{\top} X)^{-1}X^{\top} F^{\top}$, which is the same as the linear model learned without \emph{task-related} dimensions in Eq.~\eqref{eq:pseudoinverse_illposed} (the scaling constant $a$ compensates with the term at the numerator at prediction). Thus, \emph{task-related} dimensions do not affect the linear model if these are based on a tight frame, otherwise, \emph{task-related} dimensions may change the solution of the linear model.



\paragraph{Task-unrelated dimensions}
For each example $i$, we denote $\mathbf{n}_i \in\mathbb{R}^d$ a vector of $d$ \emph{task-unrelated} dimension independent from $\mathbf{x}_i \in\mathbb{R}^p$.
Thus, the new input vector to the linear model is 
$[\mathbf{x}_i^{\top} \  \mathbf{n}_i^{\top}]^{\top}$. Intuitively, if $\mathbf{n}_i^{\top}$ is randomly generated, it is expected that as the number of \emph{task-unrelated} dimensions increase there will be more overfitting. Yet, we will show several exceptions to this intuition. To illustrate the effect of these unnecessary input dimensions, we consider \emph{task-unrelated} dimensions distributed as a Gaussian, with zero mean, diagonal covariance and same variance $\sigma^2$,~\ie
$(\mathbf{n}_i)_{\ell} \sim \mathcal{N}(0,\sigma^2)$, 
$\ell = 1,\dots, d, i = 1,\dots,n$.

We assume that $p+d > n$ (\ie overparameterization), and then, the pseudo-inverse solution of the linear model in Eq.~\eqref{eq:pseudoinverse_illposed}   becomes the following:
\begin{equation}\label{eq:solution_ill_noise}
    W^+ = Y (X^{\top} X + N^{\top} N)^{-1}[X^{\top} \ N^{\top}].
\end{equation}
From this solution, we can compute the output of the linear model for a given test sample $[\mathbf{x}_{\rm ts}^\top \  \mathbf{n}_{\rm ts}^\top]^\top$ as in the following:
\begin{align} \label{W+onTest}
    W^+ [\mathbf{x}_{\rm ts}^\top \  \mathbf{n}_{\rm ts}^\top]^\top & = Y (X^\top X + N^\top N)^{-1}[X^\top \ N^\top] [\mathbf{x}_{\rm ts}^\top \  \mathbf{n}_{\rm ts}^\top]^\top \nonumber \\
    & = Y (X^\top X + N^\top N)^{-1}(X^\top \mathbf{x}_{\rm ts} +  N^\top \mathbf{n}_{\rm ts}) \nonumber \\
    & = Y(X^\top X + N^\top N)^{-1} X^\top  \mathbf{x}_{\text{ts}}   + Y(X^\top X + N^\top N)^{-1} N^\top \mathbf{n}_{\text{ts}}.
\end{align}
Note that the first term contains the minimal dimensions  and the second term the \emph{task-unrelated} dimensions. In the following, we approximate this expression to a more interpretable one.

First, we develop the last product of the second term,~\ie $N^\top \mathbf{n}_{\rm ts}$, which we write in a more explicit form:
\begin{align}
N^\top \mathbf{n}_{\rm ts} &= \left[\mathbf{n}_1^\top \mathbf{n}_{\rm ts} \ \cdots \ \mathbf{n}_n^\top \mathbf{n}_{\rm ts}         \right]^\top \\ &= \left[ \sum_{\ell = 1}^{d} (\mathbf{n}_{1})_{\ell} (\mathbf{n}_{\rm ts})_{\ell} \ \cdots \  \sum_{\ell = 1}^{d} (\mathbf{n}_{n})_{\ell} (\mathbf{n}_{\rm ts})_{\ell} \right]^\top.
\label{eq:sum_noise}
\end{align}
Recall the assumption that all $(\mathbf{n}_{i})_{\ell}$ and $ (\mathbf{n}_{\rm ts})_{\ell}$ are samples of two independent zero-mean Gaussian random variables with the same variance $\sigma^{2}$. 
For very large $d$ values, we can approximate the average realization of a random variable through its expected value (law of large numbers). 
Thus, each sum in Eq.~\eqref{eq:sum_noise} approximates  the expectation value of the product of two independent random variables, multiplied by a $d$ factor. This can be written as the product of their expectation values, which is null for zero-mean Gaussian variables,~\ie
\begin{equation}\label{eq:noise_test} \sum_{\ell = 1}^{d}  (\mathbf{n}_{i})_{\ell} (\mathbf{n}_{\rm ts})_{\ell}  \simeq 0.
\end{equation}
Therefore, for any test sample, we obtain the following approximation:
\begin{align} \label{first_simeq}
N^\top \mathbf{n}_{\rm ts} \simeq {\bf 0}
\end{align}


Similarly, for the term $N^\top N$ in  Eq.~\eqref{W+onTest}, related to \emph{task-unrelated} dimensions,  for $i\neq j$, we have: 
\begin{align} \label{forij}
\mathbf{n}_{i}^\top \mathbf{n}_{j} = \sum_{\ell = 1}^{d} (\mathbf{n}_{i})_{\ell} (\mathbf{n}_{j})_{\ell} & \simeq 
0.
\end{align}
When $i=j$, the following approximation holds: 
\begin{align} \label{forii}
\mathbf{n}_{i}^\top \mathbf{n}_{i}  = \sum_{\ell = 1}^{d} (\mathbf{n}_{i})_{\ell}^{2}  & \simeq 
d \sigma^2.
\end{align} 
 This is because the expectation value coincides with the variance of the zero-mean Gaussian variable.
Given Eqs.~\eqref{forij} and \eqref{forii}, we obtain the following approximation:
\begin{align} \label{second_simeq}
N^\top N \simeq d \sigma^2 \mathbb{I}_n.
\end{align}

Finally, the approximations in
Eqs.~\eqref{first_simeq} and \eqref{second_simeq} lead to the following approximation of the prediction of the linear network  for  large $d$:
\begin{equation}\label{eq:pinv_tikhonov}
    W^+ [\mathbf{x}_{\rm ts}^{\top} \  \mathbf{n}_{\rm ts}^{\top}]^{\top} \simeq Y (X^{\top} X + d \sigma^2 \mathbb{I}_n)^{-1} X^{\top} \mathbf{x}_{\rm ts}, 
\end{equation}
where $[\mathbf{x}_{\rm ts}^{\top} \  \mathbf{n}_{\rm ts}^{\top}]^{\top}$ is an arbitrary test sample.

We can observe by analysing Eq.~\eqref{eq:pinv_tikhonov} that $d \sigma^2 \mathbb{I}_n$ corresponds to the Tikhonov regularization term, with regularization parameter $\lambda=d\sigma^2$. 
Thus, a large number of \emph{task-unrelated} dimensions following a Gaussian distribution leads to a network trained on the dataset without \emph{task-unrelated} dimensions, biased towards small $\ell_2$ norm weights. 
This intrinsic regularization term usually will not have a positive effect in the test accuracy, as we do not have control to adjust the strength of the regularizer~\ie $d\sigma^2$ is given by the dataset. It may lead to underfitting with respect to the minimal dimensions of the dataset, or equivalently, to overfitting to the dataset with the unnecessary dimensions. Yet, this regularizer may have positive effects in some situations. In~\ref{sec:real_regression_problem}, we introduce a  regression problem with corrupted labels,~\ie $y_i = f(x_i) + \varepsilon_i$, that can benefit from the regularization effect of \emph{task-unrelated} dimensions.

\paragraph{Combining task-related and task-unrelated dimensions}
We now assume that a percentage $\nu\in[0,1]$ of unnecessary input dimensions are \emph{task-related} and the rest are  \emph{task-unrelated}. We borrow the results introduced above for each of these two types of unnecessary dimensions, and combine them. 

Let  $N = [\mathbf{n}_1 \cdots \mathbf{n}_n] \in\mathbb{R}^{d(1 - \nu)\times n}$ be the matrix of  \emph{task-unrelated} dimensions, and $T X = [T\mathbf{x}_i \cdots  T\mathbf{x}_n] \in\mathbb{R}^{d \nu \times n}$ be the \emph{task-related} dimensions, in which $T\in\mathbb{R}^{d \nu \times p}$ is a generic linear transformation. Thus, the input samples of the training set in matrix form is equal to $[X^\top \ N^\top \ (TX)^\top]^\top$, which is equivalent to the concatenation of these terms.

We can express the input matrix using the frame formalism as 
\begin{equation}
    \begin{pmatrix}
    \begin{array}{c}
        X \\ 
        N \\
        TX \\
    \end{array}
    \end{pmatrix} = 
    F \begin{pmatrix}
    \begin{array}{c}
        X \\ 
        N \\
    \end{array}
    \end{pmatrix},
\end{equation}

\begin{equation}
    \text{ where } F = 
    \begin{pmatrix}
    \begin{array}{cc}
        \mathbb{I}_{p}  &\mathbf{0}_{p\times d(1 - \nu)} \\ 
        \mathbf{0}_{d(1 - \nu) \times p}  &\mathbb{I}_{d(1 - \nu)}  \\ 
        T                      &\mathbf{0}_{d \nu\times d(1 - \nu)}  \\
    \end{array}
    \end{pmatrix}, \label{eq:frameall}
    \end{equation}
and $\mathbf{0}$ denotes the matrix with null entries and dimensions as specified in the subscripts. 
Recall that Eq.~\eqref{eq:redundant_illposed_pinv} introduced the solution of a linear model when the input is multiplied by a frame. Thus, we develop Eq.~\eqref{eq:redundant_illposed_pinv} in order to obtain a more interpretable approximation that provides insights about the effect of \emph{task-related} and \emph{unrelated} dimensions.

We first evaluate the term $F^\top F$ from Eq.~\eqref{eq:redundant_illposed_pinv} with the frame in Eq.~\ref{eq:frameall}, which yields the following: 
\begin{equation}\label{eq:FT_F}
    F^\top F = 
    \begin{pmatrix}
        \begin{array}{c|c}
        \mathbb{I}_p + T^\top T & \mathbf{0}_{p\times d(1 - \nu)} \\
        \hline
        \mathbf{0}_{d(1 - \nu)\times p} & \mathbb{I}_{d(1 - \nu)} 
        \end{array}
    \end{pmatrix},
\end{equation} 
and it is invertible. By substituting this term in the pseudo-inverse  solution of Eq.~\eqref{eq:redundant_illposed_pinv}, we obtain
\begin{align}\label{eq:pinv_third_case}
    W^+ & = Y \left(\left[X^\top \ N^\top\right]F^\top F 
        \begin{bmatrix}
        X \\
        N
        \end{bmatrix}
        \right)^{-1}
    \left[X^\top N^\top \right] F^\top \nonumber \\
       & = Y \left(X^\top \left(\mathbb{I}_p+T^\top T\right)X + N^\top N \right)^{-1} \left[X^\top \left[\mathbb{I}_p \ T^\top \right] \ N^\top \right].
\end{align}
This result does not rely on any assumption on the distribution of the \emph{task-unrelated} dimensions, neither on a specific form of the linear transformation $T$.
To get a clearer picture of the effect of  unnecessary input dimensions, in the following we make three assumptions: (i) the \emph{task-related} dimensions are generated by repeating  $k$ times the minimal set of dimensions, such that $k=(\nu d)/p$; (ii) the vector components for the \emph{task-unrelated} dimensions are drawn from zero-mean Gaussian distributions sharing the same variance value; and (iii) $d(1 - \nu)$ number of \emph{task-unrelated} dimensions is large.

Given assumption (i), $T$ is a tight frame and the quantity $T^\top T$ corresponds to the identity in $\mathbb{R}^p$. Also, the term $\mathbb{I}_p + T^\top T$ is equivalent to $(k+1)\mathbb{I}_p$ because the minimal set of dimensions are repeated $k$ times.

The solution in Eq.~\eqref{eq:pinv_third_case} applied on a new test sample $\left[\mathbf{x}_{\rm ts}^\top \ \mathbf{n}_{\rm ts}^\top \ (T\mathbf{x}_{\rm ts})^\top \right]^\top$, becomes to the following expression:
\begin{align}
    W^+ & \left[\mathbf{x}_{\rm ts}^\top \ \mathbf{n}_{\rm ts}^\top \ (T\mathbf{x}_{\rm ts})^\top \right]^\top = \nonumber \\ & Y \left((k+1)X^\top X + N^\top N \right)^{-1} \left((k+1) X^\top \mathbf{x}_{\rm ts} + N^\top \mathbf{n}_{\rm ts} \right).
\end{align}
Using assumptions (ii) and (iii), we leverage on the law of large numbers as in the \emph{task-unrelated} case introduced before. 
Here, the sum over the \emph{task-unrelated} components consists of $d(1 - \nu)$ terms, and the following two  approximations holds: $N^\top \mathbf{n}_{\rm ts}\simeq {\bf 0}$ and $N^\top N \simeq d(1 - \nu) \sigma^2 \mathbb{I}_n$. 
Thus, the prediction for a test sample corresponds to
\begin{align}\label{eq:pred_assumptions}
    W^+ & \left[\mathbf{x}_{\rm ts}^\top\ \mathbf{n}_{\rm ts}^\top \ (T\mathbf{x}_{\rm ts})^\top \right]^\top \simeq \nonumber \\
    & Y \left((k+1) X^\top X + d(1 - \nu) \sigma^2 \mathbb{I}_n \right)^{-1} 
    X^\top \mathbf{x}_{\rm ts} (k+1) = \nonumber \\
    & Y \left( X^\top X + \frac{pd(1 - \nu)}{d\nu+p} \sigma^2 \mathbb{I}_n \right)^{-1} X^\top \mathbf{x}_{\rm ts}.
\end{align}
Observe that this expression is again the pseudo-inverse solution with Tikhonov regularization as for the \emph{task-unrelated} dimensions in Eq.~\eqref{eq:pinv_tikhonov}, except that here regularization parameter is equal to  $\frac{pd(1 - \nu)}{d\nu+p} \sigma^2 $.
The regularizer term depends on the ratio between the number of \emph{task-unrelated} and \emph{task-related} dimensions, $\nu$. 
Note that as the number of \emph{task-related} dimensions increases, in the limit of $\nu$ tending to 1, the problem is equivalent to the one formulated for the minimal dimensions, with null contribution from the unnecessary input dimensions as the regularizer parameter would tend to $0$. Thus, the number of \emph{task-related} dimensions help to adjust the regularization effect of the \emph{task-unrelated} dimensions, and hence, it can alleviate its negative effects.

\subsection{Regression with corrupted output}\label{sec:real_regression_problem}

We now analyze a problem in practice that follows up from the theoretical analysis in the previous section, and that serves to show that the regularization effect of the unnecessary input dimensions can be  helpful in some cases.

We generate an overparameterized regression dataset, with $p=10$ minimal dimensions, $o=4$ output dimensions, and $d=500$ unnecessary \emph{task-unrelated} dimensions. The $p$ minimal input dimensions are drawn from $\mathcal{N}(0, \sigma_{\text{input}}^2)$, zero-mean Gaussian distribution with a shared variance value. The number of training examples across experiments is $n=7$. The corrupted output for sample $i$ is 
\begin{equation}
\mathbf{y}_i = {W}^*\mathbf{x}_i + \varepsilon, 
\end{equation}
with zero-mean Gaussian distributed random variable $\varepsilon\sim\mathcal{N}(0, \sigma_{\rm output}^2)$. 

We compare three different baselines.
We  refer to the ``\emph{with task-unrelated} solution'' as the solution computed from the samples with \emph{task-unrelated} input dimensions,~\ie $\left[\mathbf{x}_i^\top \ \mathbf{n}_i^\top\right]^\top$, using Eq.~\eqref{eq:pinv_tikhonov}. When there is no \emph{task-unrelated} dimensions, we refer to the ``\emph{without task-unrelated} solution'', as in Eq.~\eqref{eq:pseudoinverse_illposed}. Finally, we also compute the \emph{Tikhonov} regularized solution  with $\lambda=d\sigma_{\rm input}^2$ regularization parameter, on the dataset without unnecessary input dimensions.
The prediction errors for the three baselines correspond respectively to the following:
\begin{align}
    & \text{Error(\textit{with task-unrelated})} = \nonumber\\ &\left\langle \left\| Y (X^{\top} X + N^{\top} N)^{-1}[X^{\top} \ N^{\top}][\mathbf{x}_{\rm ts}^\top \ \mathbf{n}_{\rm ts}^\top]^\top - \mathbf{y}_{\rm ts}\right\|^2_2 \right \rangle, \label{eq:w_irrelevant} \\
    &\text{Error(\textit{w/o task-unrelated})} = \nonumber \\
    & \left\langle \left\| Y (X^{\top} X )^{-1}X^{\top} \mathbf{x}_{\rm ts} - \mathbf{y}_{\rm ts}\right\|^2_2 \right \rangle, \label{eq:wo_irrelevant} \\
    & \text{Error(\textit{Tikhonov})} = \nonumber \\
    & \left\langle \left\|  Y (X^{\top} X + \lambda \mathbb{I}_{n})^{-1}X^{\top} \mathbf{x}_{\rm ts} - \mathbf{y}_{\rm ts}\right\|^2_2 \right \rangle, \label{eq:tikhnov_wo_irrelevant}
\end{align}
where the sample average $\langle \ \cdot \ \rangle$ is computed on the test examples.

\begin{figure}[t]
\centering
 \includegraphics[width=1\linewidth]{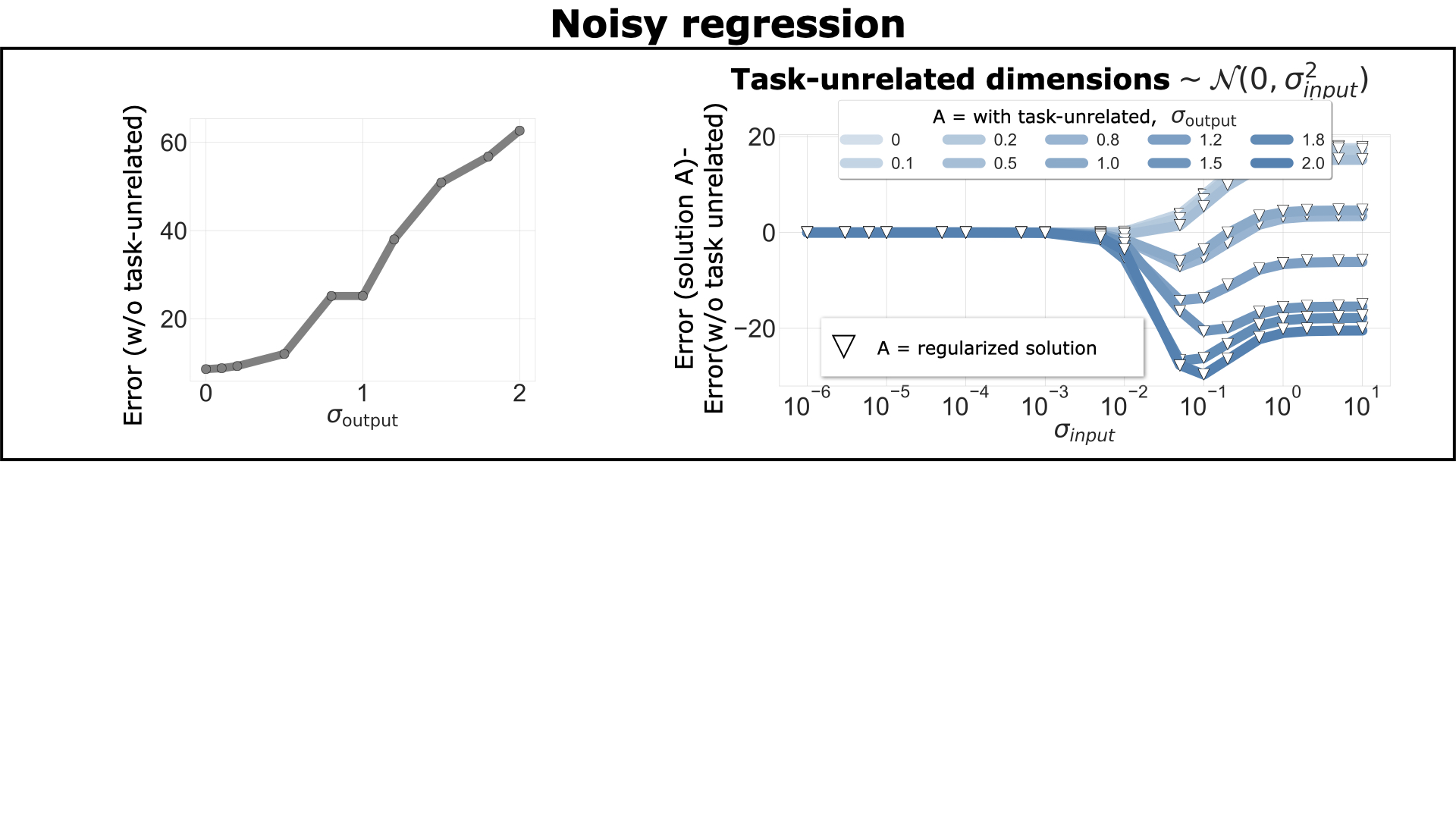}
 \vspace{-3.5cm}
\caption{Regression with corrupted output. Left: prediction error when there are no unnecessary dimensions, as function of the output corruption. Right: analysis of prediction error as function of the variance of the \emph{task-unrelated} dimensions/ regularization (with regularization value $\lambda = d \sigma_{input}^2$). On the $y$-axis, difference between test errors of either the solution ``\emph{with task-unrelated}'' or regularized, and ``\emph{without task-unrelated} solution''.}
\label{fig:noisy_regression}
\end{figure}

In Fig.~\ref{fig:noisy_regression} on the left, we report the test error computed as in Eq.~\ref{eq:wo_irrelevant}. As expected, as the  corruption in the ground-truth output increases, we observe consistently higher test errors.
On the right of Fig.~\ref{fig:noisy_regression}, we quantify the difference between Eq.~\eqref{eq:w_irrelevant} and Eq.\eqref{eq:wo_irrelevant}, as the variance of \emph{task-unrelated} dimensions increases. Each blue curve is related to a different level of output corruption. For small values of $\sigma_{\text{input}}$,  \emph{task-unrelated} dimensions do not have any effect on the test error. As $\sigma_{\rm input}$ increases, \emph{task-unrelated} dimensions harm the solution for very small output corruption, while gradually benefiting the prediction as the output corruption increases due to its regularization effect.

The white triangles on the left of Fig.~\ref{fig:noisy_regression} correspond to the difference between Eq.~\eqref{eq:w_irrelevant} and Eq.~\eqref{eq:tikhnov_wo_irrelevant}. These points follow each of the blue curves, showing that the assumptions and approximations we made are reasonable. 

Overall, these results depict the effect of  \emph{task-unrelated} dimensions, which can effectively help to improve the test accuracy when the ground-truth output is corrupted.

 \section{Supplemental Results}
\subsection{MLP on Different  Distributions of Noise}\label{app:additional_noise_non_separable}

\paragraph{Dataset and Network}
To further test the generality of previous results on data distributions that are not linearly separable, we use a mixture of Gaussians to generate non-linearly separable datasets for binary classification. Each class consists of three multivariate Gaussians of dimensions $p = 30$. As we fix a class, we generate a sample by randomly selecting, among the three, one distribution from which we draw the minimal dimensions. We evaluate the MLP with ReLU and soft-max with cross-entropy loss because among the different variants it is the only well suited to fit non-linearly separable data.
We evaluate the effect of the variance of the Gaussian distribution to generate the unnecessary dimensions, and we also evaluate two other noise distributions, namely, Gaussian noise with $\Sigma_{ii} = 1, \ \forall i$, with $\Sigma_{ij} = 0.5$, $\forall i \neq j$, and salt and pepper noise, where each vector component can assume value (0, or $u$), based on a Bernoullian distribution on $\{-1,1\}$. 
\paragraph{Results}
\begin{figure}[!ht]
\centering
    \includegraphics[width=\linewidth]{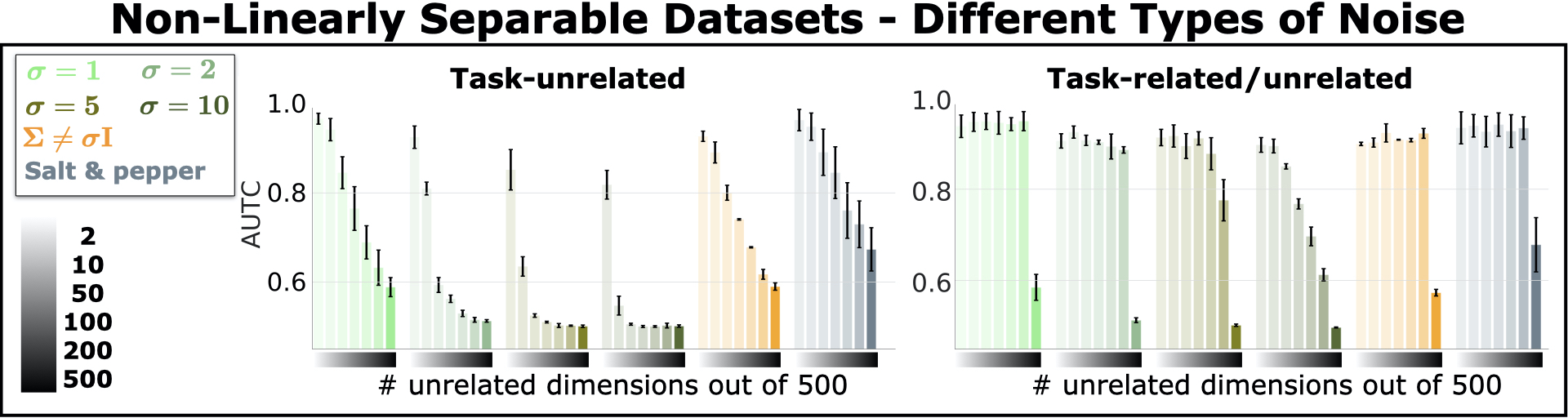}
\caption{MLP with ReLU trained with cross entropy loss on non-linearly separable binary classification dataset. The  \emph{task-unrelated} dimensions are extracted from different distributions.}
\label{fig:supplementary_different_noises}
\end{figure}

In Figure \ref{fig:supplementary_different_noises}, we show the data efficiency of MLP with ReLU trained with cross entropy loss on non-linearly separable datasets. On the left of the quadrant, we report different distributions of the \emph{task-unrelated} dimensions: Gaussian noise with different $\boldsymbol{\sigma}$ (corresponding to multiplicative factor applied on the identical covariance matrix), Gaussian noise with non-diagonal covariance matrix and salt and pepper noise. The amount of \emph{task-unrelated} dimensions is reported through the colored bar (with gradient). We observe that, similarly to previous results in the  linearly separable dataset (Figure \ref{fig:results_all}B), \emph{task-unrelated} dimensions harm data efficiency. Also, as expected, data efficiency deteriorates as the variance of Gaussian noise increases.
The combinations of \emph{task-related/unrelated} dimensions  alleviates the detrimental effect of \emph{task-unrelated} dimensions. 

\subsection{Supplementary Results on synthetic MNIST}\label{app:supplement_synthetic_mnist}
\begin{figure}[!ht]
\centering
    \includegraphics[width=\linewidth]{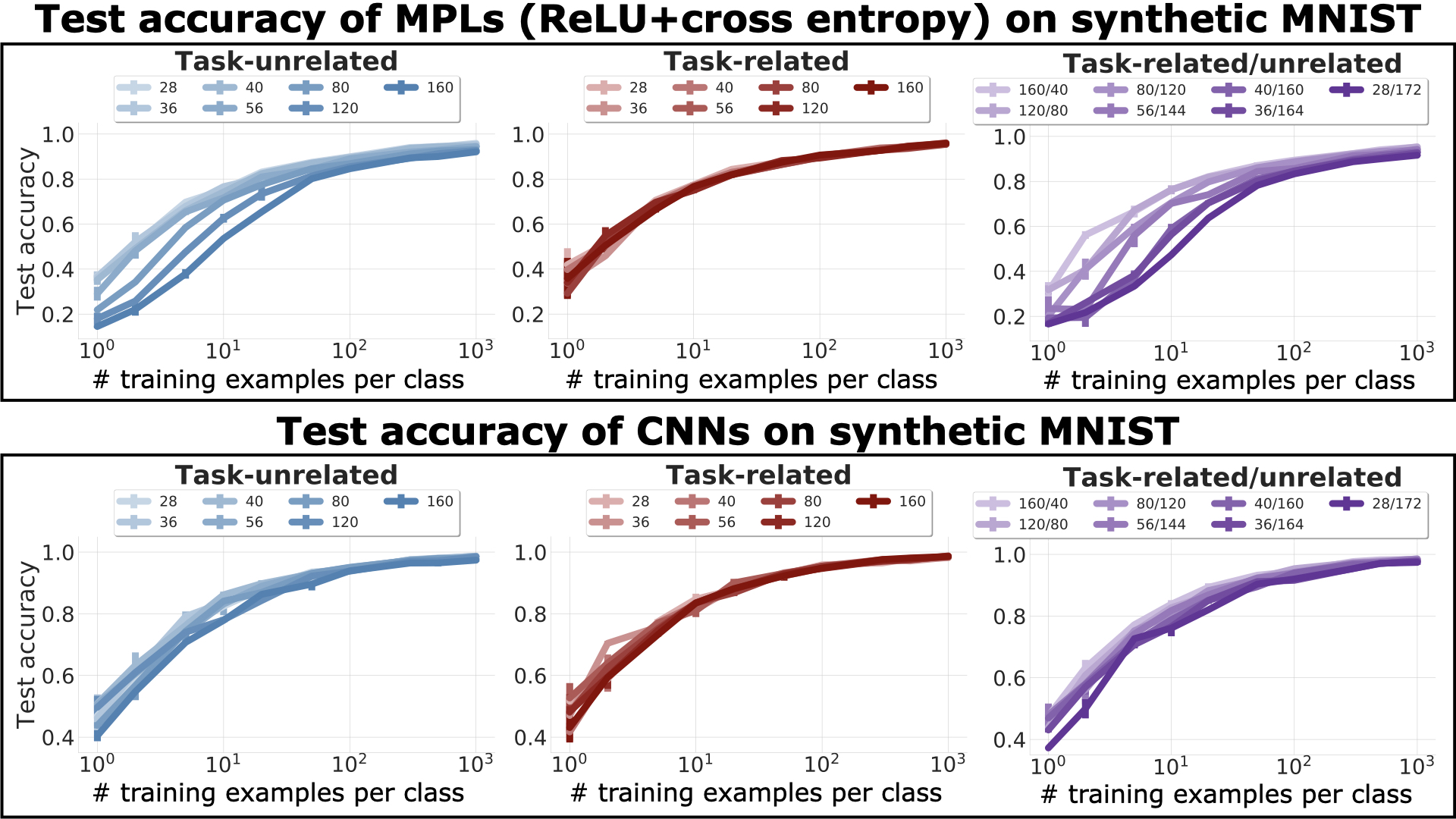}
\caption{Test accuracy for different number of training examples for the MLP with ReLU trained with cross entropy loss (top row) and CNNs (bottom row) across the three scenarios on synthetic MNIST.}
\label{fig:supplementary_synthetic_mnist}
\end{figure}

In Figure \ref{fig:supplementary_synthetic_mnist}, we report the test accuracy of the MLP and CNN, trained on synthetic MNIST, for the three types of unnecessary dimensions and the different amount of dimensions. These curves are used to compute the log-AUTC in Figure \ref{fig:results_all}C.

\subsection{Supplemental Results on natural MNIST}\label{app:supplement_natural_mnist}
\begin{figure}[!ht]
\centering
    \includegraphics[width=\linewidth]{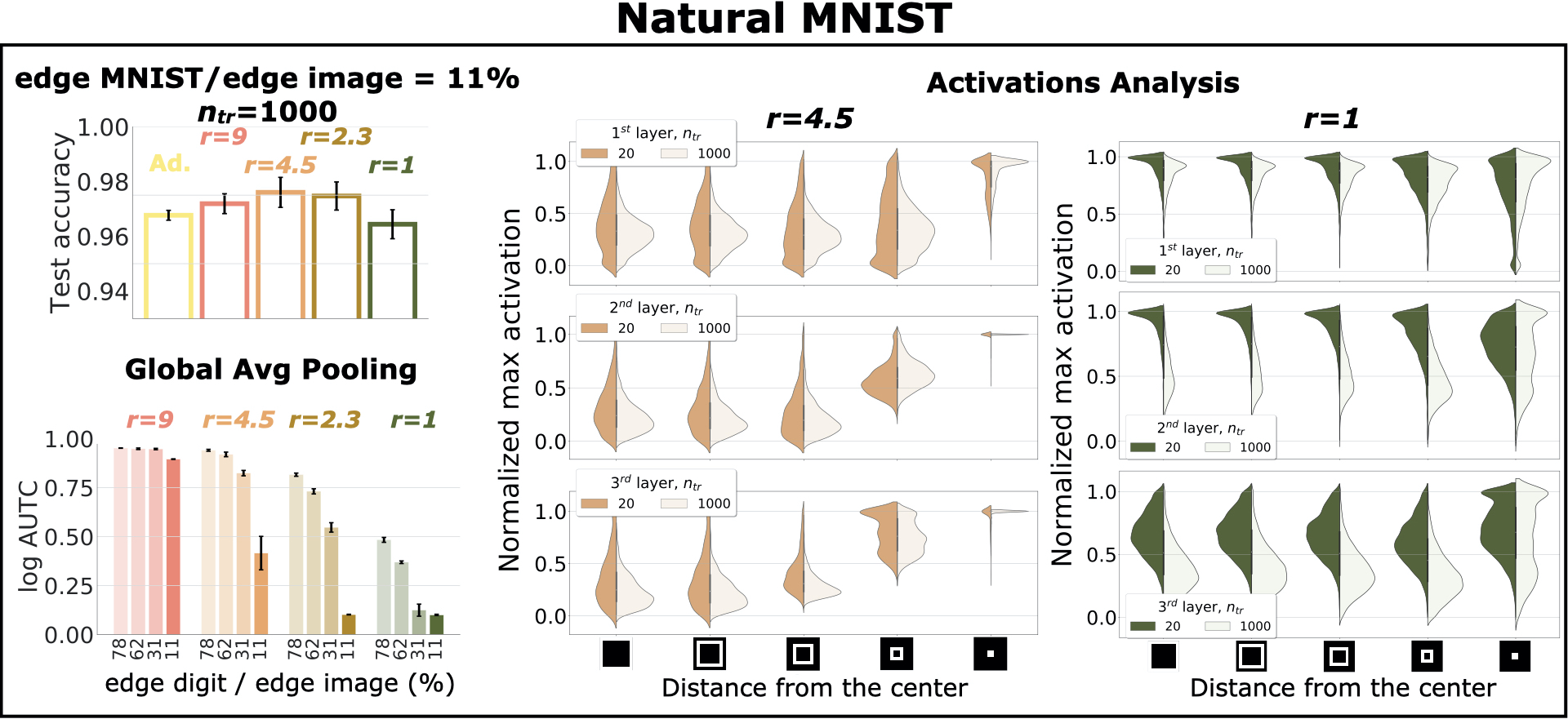}
\caption{Top left: Test accuracy for the highest amount of \emph{task-unrelated} dimensions, with the largest amount of training examples. Bottom left: log AUTC with architectures with global average pooling. On the right, normalized maximum activations for the two networks  $r=4.5$ and $r=1$ in Figure \ref{fig:results_all}D.}
\label{fig:supplementary_nat_mnist}

\end{figure}
On the top left of Figure \ref{fig:supplementary_nat_mnist}, we report the CNN's test  accuracy obtained with different filters sizes and max-pooling sizes when the amount of unnecessary dimensions, as the amount of training examples per class ($n_{tr}=1000$) are the largest we tested. All the versions of the CNN achieve almost perfect generalization, emphasizing that data efficiency's gaps due to unnecessary input dimensions are hard to observe when abundant amounts of examples are available.


On the bottom left of Figure \ref{fig:supplementary_nat_mnist}, we show the log AUTC values for networks trained on Natural MNIST with a global average pooling before the fully connected layers. The global average pooling yields fully connected layers that have the same amount of trainable parameters independently of the receptive field sizes. Results show that the CNN with the largest receptive fields size, $r=9$, leads to the highest data efficiency  among all the different $r$  we tested. CNNs with the smallest receptive field size, $r=1$, heavily suffer from the unnecessary input dimensions. Despite the reduction of free parameters caused by the global pooling, depending on the receptive field size, the data efficiency of the network could dramatically drop due to \emph{task-unrelated} input dimensions.

\paragraph{Neural Mechanisms to Discard Object's Background in CNNs}
We investigate the networks' mechanisms that facilitate discarding the unnecessary dimensions.
To do so, we provide evidence that supports the hypothesis that efficient CNNs have kernels tuned to discard the object’s background.

To compute the activations of a network, we extract at each layer the neural activity of $1000$ test samples after the non-linearity and the max-pooling (if present). For each image and at every layer, we compute the maximum activation across spatial dimensions and filters, which we use to normalize the representation of the image at that layer. We grouped the representation values at each layer in five sets, depending on their distance from the center (regions). We store the maximum normalized value for each region. 


In Figure \ref{fig:supplementary_nat_mnist}, we report the distribution of the maximum normalized values for the five regions (left to right corresponds to the furthest to the closest region to the center). We report results for the most (in orange) and least (in green) data efficient networks without global average pooling (respectively $r=4.5$ and $r=1$). Also, we compare the activations of the two networks when trained on the smallest $n_{tr} = 20$ (dark) and largest $n_{tr} = 1000$ (light) sets, for edge digit / edge image $= 11\%$.

When the network is trained with more examples (light violin plots), the neurons respond more to the object compared to the background. Yet, when the networks are trained with few examples (dark violin plots), the differences in terms of neural activity are minor between the object and its background. Note that this phenomenon is much more pronounced for the most data efficient architecture ($r=4.5$). These results suggest that the data efficiency of the network is driven by the emergence of kernels that can detect the object while not responding to the background.



\end{document}